\documentclass[letterpaper, 10 pt, conference]{ieeeconf}  

\usepackage{times}
\usepackage{xcolor}
\usepackage{graphicx}
\usepackage{amsmath}
\usepackage{amssymb}
\usepackage{booktabs}
\usepackage{url}
\usepackage{multirow}
\usepackage{url}
\usepackage[breaklinks=true,bookmarks=false]{hyperref}
\graphicspath{ {./images/} }
\usepackage{color}
\usepackage{algorithm}
\usepackage{algorithmic}
\usepackage{subfigure}
\usepackage{cases}
\usepackage[compact]{titlesec}

\IEEEoverridecommandlockouts                              

\title{\LARGE \bf
A Real-Time Fusion Framework for Long-term Visual Localization
}

\author{Yuchen Yang$^{1}$, Xudong Zhang$^{1}$, Shuang Gao$^{1}$, Jixiang Wan$^{1,2}$, \\ Yishan Ping$^{1}$, Yuyue Liu$^{3}$, Jijunnan Li$^{1}$, Yandong Guo$^{1}$
\thanks{$^{1}$All authers are with OPPO Research Institute, Shanghai, China. \{yangyuchen,  zhangxudong, gaoshuang, wanjixiang, pingyishan, lijijunnan, guoyandong\}@oppo.com.
        }%
\thanks{$^{2}$Jixiang Wan is also with Department of Automation, Shanghai JiaoTong University, Shanghai, China.}
\thanks{$^{3}$Yuyue Liu is with Wuhan University, Wuhan, China. liuyuyue98@163.com.
        }%
}

\begin{document}

\maketitle
\thispagestyle{empty}
\pagestyle{empty}

\begin{abstract}

Visual localization is a fundamental task that regresses the 6 Degree Of Freedom (6DoF) poses with image features in order to serve the high precision localization requests in many robotics applications. Degenerate conditions like motion blur, illumination changes and environment variations place great challenges in this task. Fusion with additional information, such as sequential information and Inertial Measurement Unit (IMU) inputs, would greatly assist such problems. In this paper, we present an efficient client-server visual localization architecture that fuses global and local pose estimations to realize promising precision and efficiency. We include additional geometry hints in mapping and global pose regressing modules to improve the measurement quality. A loosely coupled fusion policy is adopted to leverage the computation complexity and accuracy. We conduct the evaluations on two typical open-source benchmarks, 4Seasons and OpenLORIS. Quantitative results prove that our framework has competitive performance with respect to other state-of-the-art visual localization solutions. 

\end{abstract}

\section{INTRODUCTION}

In the robotics community, Simultaneous Localization and Mapping (SLAM) is regarded as the fundamental system for acquiring the 3D environment information and the 6DOF Poses. High-level applications such as, autonomous
driving \cite{singandhupe2019review}, Augmented Reality (AR) \cite{Castle2008Video}, Unmanned Aerial Vehicle (UAV) \cite{schmuck2017multi} and other robotic systems \cite{yousif2015overview} have high demand of precision of localization and orientation. 

Visual SLAM (VSLAM), which uses image sequences to regress the camera poses, have been developed in recent decades \cite{bodin2018slambench2}. However illumination and view-point variations, motion blur and scenery changes still have great challenges to accomplish the localization targets. Recent researches place much emphasis on front-end features in the VSLAM pipeline, aiming to extract robust features and form high-quality matches to overcome the corner cases mentioned before \cite{detone2018superpoint, revaud2019r2d2}. Such researches benefit image-based localization methods, because high performance feature tracker and keypoints matcher result in stable observations for the back-end optimization processes \cite{zhang2021reference, taira2018inloc}.

Besides, there are still some long-tail problems that exceed the limits of the visual information, which will not be solved by only images. Some researches focus on fusing Inertial Measurement Unit (IMU) with visual localization to solve such problems. Basically, fusion strategies can be separated into loosely-coupled and tightly-coupled ones. Multi-State Constraint Kalman Filter (MSCKF) \cite{mourikis2007multi} is a typical tightly-coupled solution processing propagated IMU data together with the visual features by an Extended Kalman Filter to estimate the poses. On the contrary, for loosely-coupled approaches, poses are recovered by individual pipelines before the fusion state. Generally, tightly-coupled solutions are superior in robustness and precision, yet suffer from heavy computation complexity. In order to realize the real-time application, in this work we propose a loosely-coupled architecture to balance precision and time consumption.

Considering the computational limits, we adopt the idea of distributed SLAM in the multi-robot SLAM systems to run local and global localization pipelines separately on different platforms. In multi-robot tasks, distributed SLAM systems have the extensibility to get cross-information from individual measurements, such as loop closure detection. We transfer this idea and detach the local pose estimation and the global localization with map priors. By maintaining pose communication between the server and client, we are able to accomplish the distributed visual localization system.

In this work, we propose a real-time fusion framework for long-term visual localization. Our framework is able to produce high-efficiency and high-precision localization results. Our contributions can be summarized in the following aspects:

1. We construct a refined mapping and global localization pipeline to acquire global location measurements. Some epipolar geometry hints and parallel computation ensure the accuracy and efficiency of localization.

2. We build a light-weight loosely-coupled visual localization architecture to guarantee a high calculation speed. Distributed computation strategy is adopted to achieve the server and client cooperation.

3. We design a deeply customized Pose Graph Optimization (PGO) to handle separate observations from local estimation from Visual Inertial Odomety (VIO) and global positions from global visual localization. 

4. We evaluate the final precision and run-time efficiency on both indoor and outdoor open-source datasets. We compare the proposed approach with some state-of-the-art ones to prove the performance.

\section{RELATED WORK}

\subsection{Visual Localization}
Visual localization aims at regressing the camera poses with the input images and the map prior. Basically, structure-based and retrieval-based approaches are two fundamental branches. \cite{sarlin2019coarse} constructs a complete pipeline including global descriptor retrieval, local feature matching and pose regression with PnP \cite{gao2003complete} based on 2D-3D matches corresponding to the pre-built Structure From Motion (SFM) \cite{schonberger2016structure} models. \cite{sun2021loftr} depends on Transformer networks to exceed the limits of local features in order to improve the accuracy of 2D-3D correspondences and localization performance. For retrieval-based approaches, recently researchers replace SFM model with Convolution Neural Networks (CNN) to encode the visual environments \cite{brahmbhatt2018geometry}. NetVLAD is one of the pioneers in fulfilling the localization task with CNN instead of classical ways such as BoW \cite{sivic2003video}, FV \cite{jegou2011aggregating} and VLAD \cite{jegou2010aggregating}. PoseNet \cite{kendall2017geometric} and MapNet \cite{brahmbhatt2018geometry} focus on loss function design to help the parameters better fit the environment. DSAC \cite{brachmann2018learning} and its variants indirectly predict the 3D points positions with CNNs and subsequently regress the pose with PnP. 

\subsection{Visual Inertial Odometry}
Apart from global localization, VIO systems solve relative poses in the local frames by fusing the input of images and IMU. As discussed before, MSCKF \cite{mourikis2007multi} and its variants propose a filter-based framework that treats visual features and IMU measurements fairly during the extended Kalman filter (EKF) stages. On the other hand, VINS \cite{qin2018vins} raises an highly accurate optimization-based odometry by fusing the IMU pre-integration and feature observations. ORB-SLAM3 \cite{campos2021orb} provides a multi-map system to enable the long-term place recognition and cross-map visual localization. \cite{clark2017vinet, chen2019selective, han2019deepvio} discuss the attempts to take advantage of deep learning methods to produce odometry results, but classical VIOs are still dominant for their high preicison. 

\subsection{Distributed SLAM}
Distributed SLAM is widely studied in multi-robot tasks, which relies on client-server communications to achieve multiple terminals localization results. \cite{lajoie2020door} introduces a distributed PGO architecture to combine the loop closure information from different robots, and also denies the outliers. \cite{zhang2021distributed} propose a client-server SLAM framework to keep the on-board computational and memory consumption in a low state. \cite{gouveia2014computation} assesses the network bandwidth influences on the multi-robot communication and also proves the precision gaining by the distributed architecture's help. \cite{tian2022kimera} presents a multi-robot SLAM system, that utilizes the inter-robot loop closures and metric-semantic 3D mesh model to facilitate the stable trajectory estimations.

\section{SYSTEM OVERVIEW}

\begin{figure*}
	\includegraphics[width=16cm]{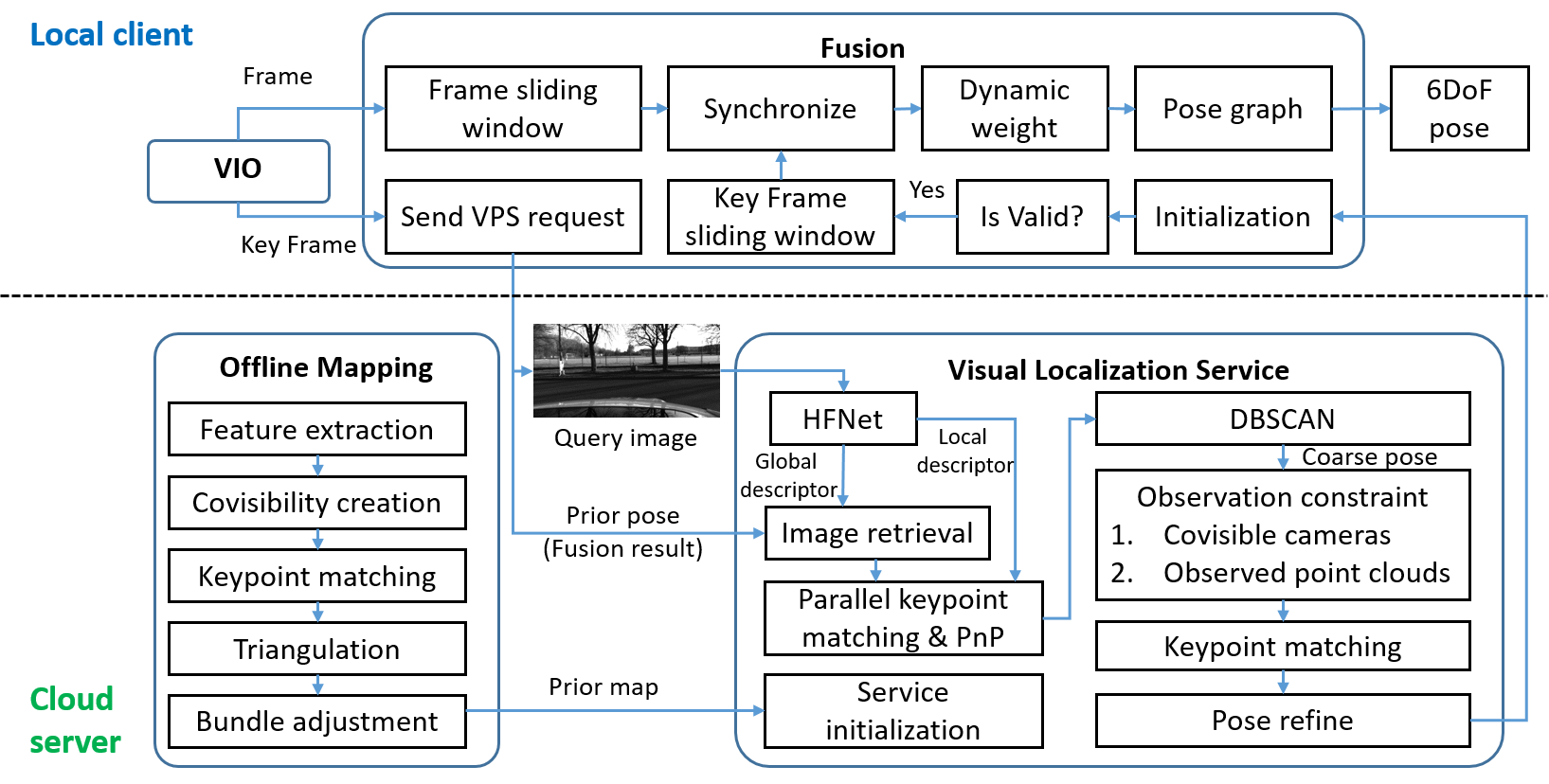}
	\centering
	\caption{
		\textbf{ System framework overview}. The system consists of four parts: Offline Mapping, Visual Localization Service, Fusion and VIO.}
	\label{fig:Framework_overview}
\end{figure*}

Our proposed client-server fusion localization framework is shown in Fig. \ref{fig:Framework_overview} including four parts. \textbf{1. Offline Mapping:} This part utilizes images with 6DoF poses to build a map as global prior information. We inherit the classical SFM pipeline and modify some typical procedures to pursue the higher performance. Specific descriptions are discussed in Sec. \ref{section: offline_mapping}.
\textbf{2. Visual Localization Service:} The service is deployed on a remote server, which is triggered by localization requests from local fusion with reference to the pre-built map and produce camera poses in the global coordinate. The details of this part will be discussed in Sec. \ref{section:vls}.
\textbf{3. VIO:} In this work, we take advantage of ORB-SLAM3 \cite{campos2021orb} to produce relative pose estimations.
\textbf{4. Fusion:} The final part fuses relative pose between frames from VIO and global localization results from visual localization service. Details of VIO and fusion will be discussed in Sec. \ref{section:vio_and_fusion}.

\section{OFFLINE MAPPING}
\label{section: offline_mapping}

In the offline mapping stage, considering the dataset formats, we believe that different strategies should be used. For sequential datasets, such as 4Seasons \cite{wenzel20204seasons} and OpenLORIS \cite{shi2020we}, image's covisible relationships can be reasoned by relative positions. Otherwise, a classical SFM problem is encountered, where image retrieval is in charge of getting the covisibility graphs. We employ HFNet \cite{sarlin2019coarse} to extract global and local features of every database images. Relatively, global features are responsible for image retrieval in localization, while local features are used to establish keypoint correspondences. Triangulation and bundle adjustment generate and refine the point clouds according to epipolar geometry. Finally, a map is created with two parts. One contains the global descriptors of mapping images. Another preserves the points cloud information observed by each mapping image. Each mapping image information is stored in one file individually. Each file includes point cloud positions, local feature descriptors and observation constraint information \cite{zhou2021retrieval}, whose notations are shown in Table. \ref{table_binary_map_structure}.

\vspace{-0.0em}
\begin{table}[h]
\caption{Binary map notation}
\label{table_binary_map_structure}
\begin{center}
\begin{tabular}{|c|c|}
\hline
Notation & Description\\
\hline
$j$           & Point cloud ID\\
\hline
$p_{j,i}^{2D}$  & 2D observation of point cloud $j$ on image $i$\\
\hline
$p_j^{3D}$  & 3D position of point cloud $j$\\
\hline
$\vec n_j$       & Observation cone normal of point cloud $j$\\
\hline
$\theta_j$  & Observation cone angle of point cloud $j$\\
\hline
$L_j$       & Maximum observation distance of point cloud $j$\\
\hline
$D_j^i$     & 2D descriptor of point cloud $j$ on image $i$\\
\hline
\end{tabular}
\end{center}
\end{table}
\vspace{-1.0em}

The maximum visible distance ${L_j}$, the mean visible direction ${\vec n_j}$ and the maximum visible angle ${\theta_j}$ can be formulated as follows:

\begin{equation} \label{eq:oc_1}
{L_j}=\mathop{max}\|p_j^{3D}-C_i\|\quad{i\in[1...N]},
\end{equation}
\begin{equation} \label{eq:oc_2}
\vec n_j=\frac{1}{N}\sum_{i=1}^{N}{\frac{{C_i-p_j^{3D}}}{{\|C_i-p_j^{3D}\|}}},
\end{equation}
\begin{equation} \label{eq:oc_3}
\theta_j=2\mathop{max}({arccos}({\vec n_j}\cdot{\frac{{C_i-p_j^{3D}}}{{\|C_i-p_j^{3D}\|}}}))\quad{i\in[1...N]},
\end{equation}

where $C_i$ denotes position of camera $i$, and point cloud $j$ is observed by $N$ cameras. The illustration of observation constraints is shown in Fig. \ref{fig:observation_constraints}. $\vec n_j$ is normalized. $\|\cdot\|$ takes magnitude of a vector.

\section{VISUAL LOCALIZATION SERVICE}
\label{section:vls}
Visual localization service (VLS) is initialized based on a prior map from Offline Mapping. After initialization, the server receives requests from fusion part and returns 6DoF localization results.

\subsection{Service initialization} 
On the service side, we occupy the scheduling strategy to maintain the memory efficiency. During the initialization, only global descriptors will be loaded to memory for the retrieval task. Subsequently, point clouds information is loaded to memory when cameras observing it is retrieved. This strategy saves memory and improves service concurrency.

\subsection{Localization}


Localization pipeline details are shown in Fig. \ref{fig:Framework_overview}. Corresponding to the mapping stage, global descriptors are used to retrieve images similar with query image based on the prior poses in the database. In detail, if $k$ reference image is needed, we retrieve top $10k$ and find reference images that distance between it and prior pose is in a threshold. If result is less than $k$, which means that prior pose is probably not accurate enough, we fill $k$ reference images with images of high scores according to retrieval score rank. If there is no prior pose, image retrieval is based on retrieval score rank. As for reference images which are close to each other, only reference image with highest retrieval score is kept, which enlarges image retrieval range.

After obtaining $k$ reference image candidates which are covisible to query image, parallel keypoint matching between local descriptors of query image and that of point clouds observed by each candidate gives 2D-3D matches with low time consumption. Parallel PnP uses $k$ groups of 2D-3D matches to provide $k$ candidate poses. In these poses, correct ones are close to each other and DBSCAN is used to filter out outlier poses. After that, we use all 2D-3D matches of inlier poses to refine the pose. Duplicate point cloud ID is ignored.

\begin{figure}
    \centering
    \subfigure[Calculate observation constraints]{
    \includegraphics[width=3.5cm]{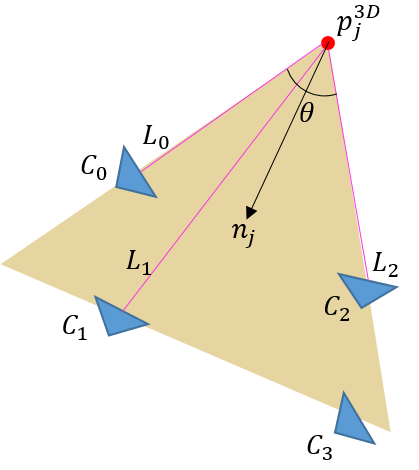}
    }
    \subfigure[Filter out cameras]{
    \includegraphics[width=3.5cm]{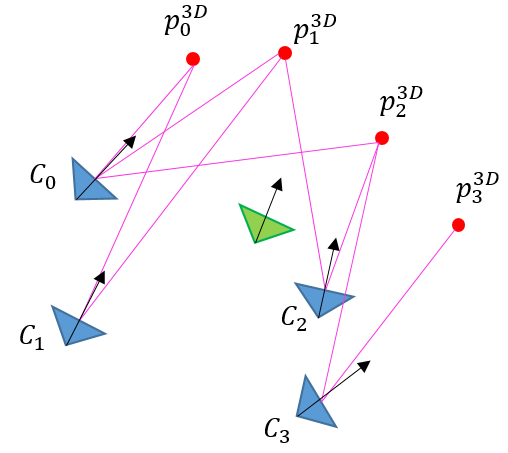}
    }
    \subfigure[Filter out point clouds]{
    \includegraphics[width=3.5cm]{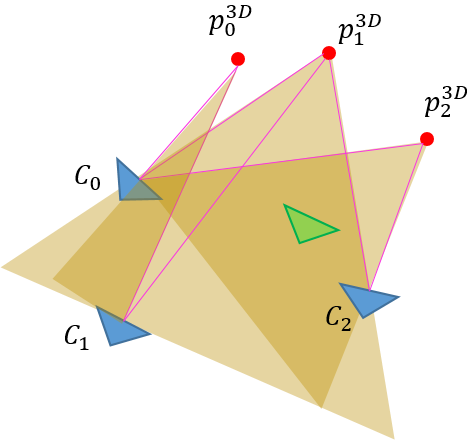}
    }
    \subfigure[Selected point clouds]{
    \includegraphics[width=3.5cm]{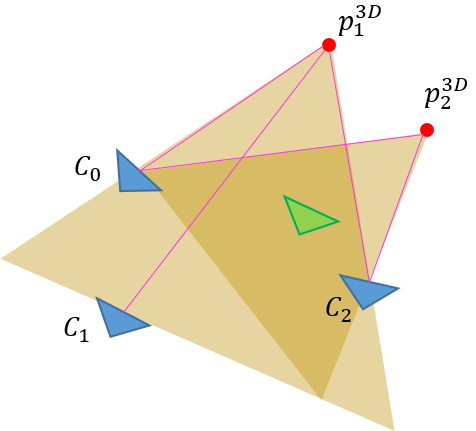}
    }
    \caption{\textbf{Illustration of Observation Constraints. }Blue triangle represents reference cameras. Red points represents point clouds. Pink lines represent observation between cameras and point clouds. Green triangle represents coarse localization result. (a). Yellow triangle represents visible cone of point cloud $p_j^{3D}$. (b). We firstly filter reference cameras co-visible with coarse camera pose using distance and angle between normal vectors of coarse camera pose and reference cameras. Camera $C_3$ is filtered out. (c). We filter out point clouds whose visible cone not covering coarse camera pose. Point cloud $p_0^{3D}$ is filtered out. (d). Finally we obtain point clouds that possibly observed by coarse camera pose.}
    \label{fig:observation_constraints}
\end{figure}

In observation constraint part, we want to find more 2D-3D matches to improve localization accuracy. As for point cloud retrieval, \cite{qin2021light} uses octree to store point cloud and realize fast point cloud retrieval and update. In order to save memory and improve retrieval speed, we firstly retrieve reference cameras that are close to coarse pose and have similar normal vectors. Then we retrieve in point clouds observed by these covisible cameras to obtain point clouds satisfying observation constraints:

1. ${\|p_c - p_j^{3D}\| < L_j + \Delta L}$,

2. ${2{arccos}({\vec n_j}\cdot{\frac{{p_c-p_j^{3D}}}{{\|p_c-p_j^{3D}\|}}}) < \theta_j + \Delta \theta}$

where ${p_c}$ is coarse camera position and $p_j$ is point cloud $j$ position. $L_j$ is maximum observation distance of point cloud $j$. $\theta_j$ is observation cone angle of point cloud $j$. $\Delta L$ and $\Delta \theta$ enlarge searching range. An illustration of observation constraint is shown in Fig. \ref{fig:observation_constraints}.

Finally, we match keypoint descriptors of query image and point clouds' descriptors again and refine the coarse pose.

\section{VIO and Fusion}
\label{section:vio_and_fusion}

VIO and fusion are on the local client. VIO provides real-time relative pose between images. Based on a pose graph, fusion uses relative poses from VIO and visual localization result to calculate real-time camera pose.

\subsection{Framework}
As shown in Fig. \ref{fig:Framework_overview}, VIO common frames and key frames are the input of fusion. 

As for key frames, we send visual localization request to server. A request includes key frame image, timestamp and prior pose. Fusion needs initialization first. If fusion is initialized, prior pose is latest fusion result, while if fusion is not initialized, there is no prior pose.
After initialization, we judge whether this key frame is valid according to its visual localization result. Valid key frames are imported into key frame sliding window.

As for common frames, we also maintain a sliding window. We synchronize timestamps between sliding windows and calculate dynamic weights for edges in PGO. Finally, PGO outputs 6DOF fusion result.

\subsection{Initialization}
Considering similar scenes may lead to incorrect visual localization result, a multi-frame initialization is used to provide robust initial transformations between VIO and VLS. For every key frame $i$, we calculate transformation between VIO and VLS coordinate $\hat{T}_{d,i}$:
\begin{equation}\label{eq:valid_3}
    {\hat{T}}_{d,i} = {\hat{T}}_{o,i}^{l,i} = {\hat{T}}_{i}^{l}{\hat{T}}_{o}^{i},
\end{equation}
where ${\hat{T}}_{i}^{o}$ and ${\hat{T}}_{i}^{l}$ indicate $4\times4$ transformation matrix from key frame $i$ to VIO and VLS coordinate separately. ${\hat{(\cdot)}}$ denotes a noisy measurement.

K-Means algorithm clusters ${\hat{T}}_{d,i}$ of success visual localization result. When a cluster has enough visual localization results, initialization succeed and key frames in this cluster is loaded to key frame sliding window.

When visual localization keep failing for over 20 key frames, we believe that old drift is unapplicable and conduct a re-initialization, which is same as initialization.

\subsection{Key frame validation}
Considering visual localization result may zigzag while VIO trajectory is smooth, we judge whether visual localization is valid according to adjacent key frame VIO pose and local drift distribution between VIO coordinate and visual localization coordinate:

\begin{equation}\label{eq:valid_1}
    \|[{\hat{T}}_{o}^{i+1}{\hat{T}}_{i}^{o}]_p - [{\hat{T}}_{l}^{i+1}{\hat{T}}_{i}^{l}]_p\| < D_v
\end{equation}
\begin{equation}\label{eq:valid_2}
    [{\hat{q}}_{o}^{i}{\hat{q}}_{i+1}^{o}{\hat{q}}_{l}^{i+1}{\hat{q}}_{i}^{l}]_{degree} < R_v
\end{equation}
\begin{equation}\label{eq:valid_4}
    {\hat{T}}_{d,i}\sim{N(T_{d}, \sigma^2)}
\end{equation}
\begin{equation}\label{eq:valid_5}
    [{\hat{T}}_{d,i}T_{d,i}^{-1}]_{p,q} < 3\sigma
\end{equation}

$[\cdot]_p$ takes position part of transformation matrix. $[\cdot]_{degree}$ takes rotation degree of a quaternion. 
$\hat{q}_{o}^i$ and $\hat{q}_{l}^i$ indicates quaternion of rotation from VIO and VLS coordinate to frame $i$ separately. $[\cdot]_{p,q}$ takes position and quaternion of a $4\times4$ transformation matrix. $D_v$ and $R_v$ are distance and rotation valid thresholds. We believe that trajectory of VIO and VLS in a short period is similar. So we suppose that $\hat{T}_{d,i}$ follows a normal distribution(Eq. \ref{eq:valid_4}) and a new valid $\hat{T}_{d,i}$ is supposed to satisfy $3\sigma$ rule in Eq. \ref{eq:valid_5}.
After all, a key frame is valid when it satisfy Eq. \ref{eq:valid_1}, Eq. \ref{eq:valid_2} and Eq. \ref{eq:valid_5}.

\subsection{Sliding window}
We maintain two local sliding window for common frames and key frames separately. Key frame sliding window has fixed frame size. Common frames sliding window shares same timestamp range with key frame sliding window and will drop old frames when key frame sliding window is updated. Every common frame is imported into common frame sliding window and optimized in pose graph.

Considering key frame rate depends on VIO while increasing common frames may increase PGO time consumption, latest common frame may fall behind latest key frame. Synchronization aligns key frame and common frames and if common frame fall behind, we only give initial fusion state to frames and skip optimization. Initial fusion state of new frames is given by Eq. \ref{eq:state_init}:
\begin{equation}\label{eq:state_init}
    T_{i}^f = T_d \hat{T}_{i}^{o},
\end{equation}
where $T_d$ is given by Eq. \ref{eq:valid_4}. $T_{i}^f$ denotes camera pose of frame $i$ in fusion coordinate.

\begin{figure}
    \centering
    \includegraphics[width=7cm]{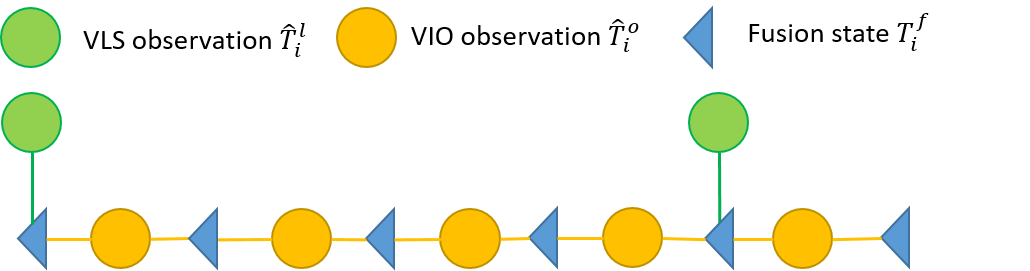}
    \caption{\textbf{Illustration of pose graph. }Pose graph includes VIO edges and VLS edges. Not every camera state has a VLS edge because visual localization is for key frames and VLS result may fail or invalid because of challenging scenes. VIO edges connects every two adjacent frames.}
    \label{fig:pose_graph}
\end{figure}

\subsection{Pose graph}
In pose graph, as illustrated in Fig. \ref{fig:pose_graph}, it has VIO edges and VLS edges. Optimization target is fusion state of all frames in the sliding window. Considering position residual is usually larger than rotation, to balance weight between position loss and rotation loss, pose graph is divided into two steps. In the first step, position and rotation are both optimized. In the second step, only rotation is optimized. First step optimization is described in Eq. \ref{eq:PGO_1}.

\begin{equation} \label{eq:PGO_1}
\begin{split}
 \mathop{min}\limits_{T_{0}^f,...T_{n}^f} \left\{  {\sum\limits_{i=0}^{n-1}{\alpha\|{r_{o}(T_{i}^f,T_{i+1}^f,{\hat{z}}_{i,i+1}^{o})}\|}^2} \right. \\  \left. +\sum\limits_{i=0}^{n-1}{{\beta}_i\|{h(i)r_{l}(T_{i}^f,{\hat{z}}_{i}^{l})}\|}^2 \right\}
\end{split}
\end{equation}

\begin{equation}
    \hat{z}_{i,i+1}^{o} = [{\hat{T}_{o}^{i+1}} \hat{T}_i^{o}]_{p,q} = 
    \begin{bmatrix}
        \delta{\hat{p}}_{i,i+1}^{o} \\
        \delta{\hat{q}}_{i,i+1}^{o}
    \end{bmatrix}
\end{equation}

\begin{equation}
    {\hat{z}}_{i}^{l} = [{\hat{T}_{i}^{l}}]_{p,q} = 
    \begin{bmatrix}
        {\hat{p}}_{i}^{l} \\
        {\hat{q}}_{i}^{l}
    \end{bmatrix}
\end{equation}

\begin{equation}
    h(i)= \begin{cases}
    1,\quad &\text{frame $i$ has VLS edge} \\
    0,\quad &\text{frame $i$ has no VLS edge}
    \end{cases} 
\end{equation}

$r_{o}(\cdot)$ is residuals of VIO edges. $r_{l}(\cdot)$ denotes residuals of VLS edges. Residuals are defined in Eq. \ref{eq:res_vio} and Eq. \ref{eq:res_vls}. $\hat{z}_{i,i+1}^{o}$ denotes VIO observation, which is relative pose from frame $i$ to $i+1$. $\hat{z}_{i}^{l}$ denotes VLS observation, which is camera pose of frame $i$ in VLS coordinate. $\alpha$ and ${\beta}_i$ denote weights of two types of residuals. $\alpha$ is a fixed value while ${\beta}_i$ is dynamic according to VLS error, which is defined if Eq. \ref{eq:drift_ransac_4}.

\begin{equation}\label{eq:res_vio}
    r_{o}(T_{i}^f,T_{i+1}^f,\hat{z}_{i,i+1}^{o}) = 
    \begin{bmatrix}
        w_t(R_f^i(p_{i+1}^f - p_{i}^f) - \delta{\hat{p}}_{i,i+1}^{o} \\
        w_q(\left[{q_{i}^f}^{-1} {q_{i+1}^f} \delta{\hat{q}}_{i,i+1}^{o}\right]_{xyz})
    \end{bmatrix}
\end{equation}

\begin{equation}\label{eq:res_vls}
    r_{l}(T_{i}^f,\hat{z}_{i}^{l}) = 
    \begin{bmatrix}
        w_t(p_{i}^f - \hat{p}_{i}^{l}) \\
        w_q(\left[{{q_{i+1}^f}^{-1}}{\hat{q}_{i}^{l}}\right]_{xyz})
    \end{bmatrix}
\end{equation}

In Eq. \ref{eq:res_vio} and Eq. \ref{eq:res_vls}, $[\cdot]_{xyz}$ takes first three parts of quaternion. $w_t$ and $w_q$ are weights of translation and rotation in residual calculation. Considering quaternion vector is normalized while translation is not, rotation weight should be higher than translation. In our experiment, $w_q=10w_t$.

Second step optimization only includes rotation, whose definition is similar with first but only include rotation residuals in VIO and VLS edges.

\begin{figure}
    \centering
    \includegraphics[width=6cm]{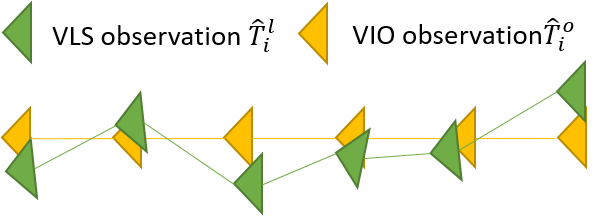}
    \caption{\textbf{Illustration of drift RANSAC. }Drift RANSAC randomly samples local key frames and calculate errors of each set. Then we estimates ${T_{o}^{l}}^{\prime}$ between VIO and VLS coordinates of least error among ${T_{o,k}^{l}}$. Finally we calculate error and weight of each VLS edge.}
    \label{fig:drift_ransac}
\end{figure}

As discussed before,  VLS results are not smooth. VLS edge with low error should be given a higher weight. Although we estimate normal distribution of local drift in Eq. \ref{eq:valid_4}, error of each VLS is still unknown. VLS edge with large error always contributes to normal distribution and needs to be eliminated. A drift RANSAC calculates dynamic weight of VLS edges, as shown in Fig. \ref{fig:drift_ransac}. We randomly sample 100 sets from $N$ key frames and each set contains 4 key frames. Umeyama method \cite{umeyama1991least} is utilized to estimate $T_{o,k}^{l}$ of set $k$. The error of each set is defined as:

\begin{equation}\label{eq:drift_ransac_1}
    T_{o,k}^{l} = u(S_k),
\end{equation}
\begin{equation}\label{eq:drift_ransac_2}
    e_k = \sum\limits_{i=1}^{N}{e_k^i} = \sum\limits_{i=1}^{N}{\|[T_{o,k}^{l}T_{i}^{o}]_p - [T_{i}^{l}]_{p}\|},
\end{equation}
where $S_k$ is set $k$ in RANSAC. $u(\cdot)$ is Umeyama method. $e_k$ denotes error of set $k$. $[\cdot]_{p}$ takes position part of a 4x4 transformation matrix. Eq. \ref{eq:drift_ransac_2} calculate error of set $k$. We select set with minimum $e_k$ and denote its $T_{o,k}^{l}$ as ${T_{o}^{l}}^{\prime}$. Then we calculate dynamic weight $ {\beta}_i$ of each VLS edge:

\begin{equation}\label{eq:drift_ransac_4}
    {\beta}_i = \frac{1}{\|[{T_{o}^{l}}^{\prime}T_{i}^{o}]_p - [T_{i}^{l}]_{p}\| + 1}
\end{equation}

In the pose graph of a sliding window, considering that nodes at two sides of sliding window have fewer edge constraints and is unstable during optimization, pose of frame in the middle of sliding window is used as final input. As for prior pose for VLS, we still utilize the latest frame in the sliding window.

\section{EXPERIMENTAL RESULTS}
\label{section:experimental_results}

We tested our system on both outdoor and indoor datasets. VIO used in our system is stereo-inertial ORB-SLAM3\cite{campos2021orb}. We change parameters in visual-IMU initialization part to fasten initialization. VLS is deployed on a remote server with NVIDIA T4 and Intel Xeon Gold 5218 CPU(2.30GHz). VIO and fusion part is deployed on a desktop PC with Intel Xeon CPU E3-1240 v6(3.70GHz).

\subsection{Datasets}
Our method aims at real-time and long-term localization in sequential changing scenes, so we choose 4Seasons(outdoor) \cite{wenzel20204seasons} and OpenLORIS(indoor) \cite{shi2020we}. 4Seasons dataset includes stereo images collected in different seasons and weather. It also covers different scenes such as countryside, town and parking garage. Ground truth poses are provided by fusion of RTK-GNSS and direct stereo visual-inertial odometry. OpenLORIS include color, depth and fisheye stereo images with dynamic objects, illumination variety and scene changes. It also includes five different indoor scenes: cafe, corridor, office, market and home. For office scene, ground truth is obtained from a Motion Capture System. For other scenes, it is provided by a 2D laser SLAM.

We evaluate both absolute and relative localization accuracy of our method. As for absolute localization accuracy, we choose a state-of-the-art visual localization toolbox hloc \cite{sarlin2019coarse} which uses NetVLAD \cite{arandjelovic2016netvlad}, SuperPoint \cite{detone2018superpoint} and SuperGlue \cite{sarlin2020superglue} as image retrieval, feature point and matching method. We evaluate the percentage of query images within three different thresholds of rotation and translation error to quantify visual localization accuracy. We choose ORB-SLAM3 \cite{campos2021orb} to compare RMSE of absolute trajectory error(ATE) of position with our method.

\subsection{Visual localization accuracy}
In Table. \ref{table:loc_result_4seasons} and \ref{table:loc_result_openloris}, the second column indicates train and test set. $1/2$ means that mapping is based on set 1 and localization test utilizes set 2. ORB-SLAM3 trajectory is aligned to ground-truth with 6DOF. Ours and hloc results are in ground-truth coordinate and are not aligned.

\begin{table}[htbp]
\caption{Visual localization accuracy on 4Seasons dataset.}
\label{table:loc_result_4seasons}
\centering
\scalebox{0.68}{
\setlength{\tabcolsep}{1mm}{
\begin{tabular}{cccccc}
\hline
\multicolumn{1}{c}{\textbf{Scene}} & \multicolumn{1}{c}{\textbf{\begin{tabular}[c]{@{}c@{}}Train/\\Test\end{tabular}}} & \multicolumn{1}{c}{\textbf{\begin{tabular}[c]{@{}c@{}}NetVLAD\cite{arandjelovic2016netvlad} top20+\\ SuperPoint\cite{detone2018superpoint}+\\SuperGlue\cite{sarlin2020superglue}\end{tabular}}} & \multicolumn{1}{c}{\textbf{Ours}} & \multicolumn{1}{c}{\textbf{Ours}} & \multicolumn{1}{c}{\textbf{ORB-SLAM3}} \\ \hline
                                   &                 & \multicolumn{2}{c}{(0.25m, 2°) / (0.5m, 5°) / (5m, 10°)} & \multicolumn{2}{c}{ATE RMSE(m)}                     \\ \hline
BusinessCampus                     & 1/2 & \textbf{92.90\%} 97.41\% 98.98\%  & 83.94\% \textbf{98.04\% 99.19\%}         & \textbf{0.191} & 9.878     \\ 
                                   & 1/3 & 1.87\% 94.72\% 96.56\%            & \textbf{4.21\% 96.53\% 99.53\%}          & \textbf{0.218} & 7.271     \\ \hline
CityLoop                           & 2/1 & \textbf{32.69\%} 72.67\% 85.77\%  & 30.02\% \textbf{76.44\% 95.16\%}         & \textbf{1.586} & 241.682   \\ 
                                   & 2/3 & \textbf{58.21\%} 75.13\% 81.40\%  & 57.35\% \textbf{85.13\% 94.04\%}         & \textbf{1.543} & 111.508   \\ \hline
Countryside                        & 3/1 & \textbf{35.70\%} 50.13\% 71.36\%  & 34.90\% \textbf{67.88\% 98.03\%}         & \textbf{0.799} & 120.666   \\ 
                                   & 3/2 & \textbf{34.51\%} 48.96\% 72.37\%  & 30.27\% \textbf{55.14\% 97.75\%}         & \textbf{1.183} & 68.651    \\ 
                                   & 3/4 & 21.86\% 34.35\% 60.71\%           & \textbf{24.14\% 51.67\% 96.20\%}         & \textbf{0.902} & 89.135    \\ \hline
OldTown                            & 2/1 & \textbf{78.71\%} 92.31\% 96.26\%  & 74.86\% \textbf{93.47\% 99.94\%}         & \textbf{0.397} & 41.399    \\ 
                                   & 2/3 & \textbf{29.95\%} 96.60\% 98.91\%  & 26.51\% \textbf{96.87\% 100.00\%}        & \textbf{0.169} & 26.315    \\ \hline
Neighborhood                       & 5/1 & 84.43\% 91.20\% 92.00\%           & 83.22\% \textbf{92.38\% 100.00\%}        & \textbf{0.645} & 2.631     \\
                                   & 5/2 & 0.73\% \textbf{85.14\%} 88.03\%   & \textbf{2.64\%} 84.97\% \textbf{94.33\%} & \textbf{1.592} & 3.638     \\
                                   & 5/3 & 0.39\% 83.74\% 86.20\%            & \textbf{2.02\% 84.54\% 97.87\%}          & \textbf{1.429} & 3.511     \\
                                   & 5/4 & \textbf{56.95\%} 90.18\% 91.37\%  & 56.63\% \textbf{90.64\% 97.83\%}         & \textbf{1.278} & 3.264     \\
                                   & 5/6 & \textbf{64.03\%} 71.64\% 87.29\%  & 61.38\% \textbf{72.71\% 99.40\%}         & \textbf{0.591} & 5.045     \\
                                   & 5/7 & \textbf{83.63\%} 87.72\% 90.65\%  & 81.29\% \textbf{91.51\% 95.56\%}         & \textbf{0.207} & 5.717     \\ \hline
ParkingGarage                      & 2/1 & \textbf{3.78\% 76.23\% 99.94\%}   & 3.55\% 75.77\% 98.91\%                   & \textbf{0.409} & 3.056     \\ 
                                   & 2/3 & 41.13\% 74.88\% 98.93\%           & \textbf{42.21\% 76.60\% 100.00\%}        & \textbf{0.418} & 2.410     \\ \hline
OfficeLoop                         & 1/2 & \textbf{99.00\% 99.84\% 100.00\%} & 97.84\% 99.24\% 99.58\%                  & \textbf{0.092} & 8.623     \\ 
                                   & 1/3 & 0.00\% 90.49\% 94.59\%            & 0.00\% \textbf{94.20\% 99.47\%}          & \textbf{0.229} & 13.415    \\ 
                                   & 1/4 & 2.44\% 65.34\% 70.91\%            & \textbf{6.20\% 68.05\% 93.53\%}          & \textbf{2.526} & 12.000    \\ 
                                   & 1/5 & \textbf{78.26\%} 94.17\% 96.12\%  & 76.09\% \textbf{97.94\% 98.97\%}         & \textbf{0.158} & 10.553    \\ 
                                   & 1/6 & 89.21\% 95.83\% 97.43\%           & \textbf{90.71\% 99.00\% 99.45\%}         & \textbf{0.123} & 12.090    \\ \hline

\end{tabular}
}
}
\end{table}

In OpenLORIS dataset, market scene is unused because three trajectories have few over lapping. Office 3 scene is too short and has few shared scene with train set and is also unused. Fisheye stereo images are used for VIO and color images are used for mapping and visual localization.

\begin{table}[htbp]
\caption{Visual localization accuracy on OpenLORIS dataset.}
\label{table:loc_result_openloris}
\centering
\scalebox{0.7}{
\begin{tabular}{cccccc}
\hline
\multicolumn{1}{c}{\textbf{Scene}} & \multicolumn{1}{c}{\textbf{\begin{tabular}[c]{@{}c@{}}Train/\\Test\end{tabular}}} & \multicolumn{1}{c}{\textbf{\begin{tabular}[c]{@{}c@{}}NetVLAD\cite{arandjelovic2016netvlad} top20+\\ SuperPoint\cite{detone2018superpoint}+\\SuperGlue\cite{sarlin2020superglue}\end{tabular}}} & \multicolumn{1}{c}{\textbf{Ours}} & \multicolumn{1}{c}{\textbf{Ours}} & \multicolumn{1}{c}{\textbf{ORB-SLAM3}} \\ \hline
                                   &     & \multicolumn{2}{c}{(0.1m, 1°) / (0.25m, 2°) / (1m, 5°)} & \multicolumn{2}{c}{ATE RMSE(m)}                    \\ \hline
cafe                               & 2/1 & \textbf{67.10\% 78.51\%} 83.26\%  & 56.99\% 75.19\% \textbf{87.83\%}             & 0.147 & \textbf{0.094}    \\ \hline
corridor                           & 1/2 & \textbf{46.55\%} 56.27\% 59.61\%  & 45.79\% \textbf{87.84\% 97.76\%}             & \textbf{0.131} & 0.444    \\ 
                                   & 1/3 & 24.02\% 29.17\% 29.93\%           & \textbf{26.52\% 48.38\% 85.76\%}             & \textbf{0.457} & 0.670    \\ 
                                   & 1/4 & 37.47\% 40.38\% 43.03\%           & \textbf{47.33\% 64.37\% 80.31\%}             & \textbf{0.173} & 0.427    \\ 
                                   & 1/5 & \textbf{70.69\%} 84.03\% 89.44\%  & 61.40\% \textbf{85.28\% 94.29\%}             & \textbf{0.133} & 0.372    \\ \hline
home                               & 1/2 & \textbf{4.22\%}  28.90\% 50.45\%  & 2.08\% \textbf{51.46\% 92.66\%}              & \textbf{0.279} & 0.356    \\ 
                                   & 1/3 & \textbf{15.87\% 33.94\%} 49.13\%  & 6.45\% 22.53\% \textbf{95.56\%}              & \textbf{0.345} & 0.366    \\ 
                                   & 1/4 & 11.67\% 35.23\% 47.55\%           & \textbf{11.99\% 43.60\% 94.50\%}             & \textbf{0.297} & 0.319    \\ 
                                   & 1/5 & \textbf{11.21\%} 37.63\% 68.43\%  & 8.51\% \textbf{41.49\% 86.60\%}              & \textbf{0.289} & 0.319    \\ \hline
office                             & 5/1 & 38.32\% 54.51\% 60.57\%           & \textbf{42.59\% 69.75\% 75.31\%}             & \textbf{0.179} & 0.063    \\
                                   & 5/2 & 63.29\% 79.76\% 94.77\%           & \textbf{68.89\% 90.78\% 91.33\%}             & \textbf{0.046} & 0.084    \\
                                   & 5/4 & \textbf{45.75\%} 59.20\% 69.66\%  & 41.49\% \textbf{69.89\% 91.15\%}             & 0.232 & \textbf{0.153}    \\
                                   & 5/6 & \textbf{26.76\%} 42.96\% 54.07\%  & 21.76\% \textbf{50.65\% 57.31\%}             & 0.254 & \textbf{0.059}    \\
                                   & 5/7 & 8.33\%  \textbf{38.21\%} 70.73\%  & \textbf{10.00\%} 34.04\% \textbf{93.33\%}    & 0.223 & \textbf{0.051}    \\ \hline
\end{tabular}
}
\end{table}

\begin{figure}
    \centering
    \includegraphics[width=7cm]{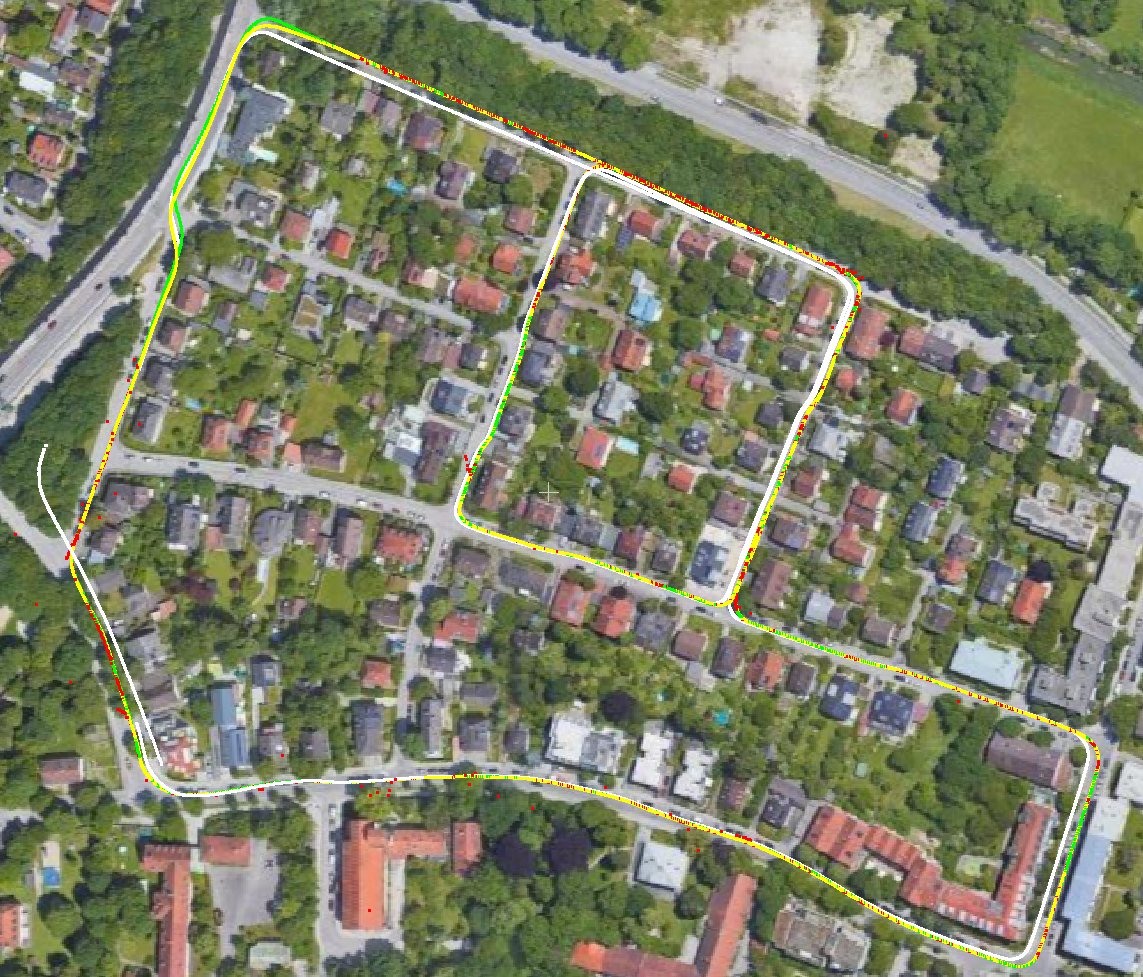}
    \caption{
    Illustration of localization result. Yellow line is our result. White line is ORB-SLAM3 result. Red points are hloc result. Green line is ground truth. Map image is from Google Map.
    }
    \label{fig:neighborhood_result}
\end{figure}

As shown in Table. \ref{table:loc_result_4seasons}, our fusion framework realizes accurate and stable localization in various scenes. Especially in countryside scene, which is vegetarian scenes and challenging for feature matching, our method realizes higher success rate by VIO. In neighborhood scene, test sets have parts that train set did not cover. Localization results of different methods are shown in Fig. \ref{fig:neighborhood_result}. Our method is able to localize for all images. However hloc fails in locations out of map but performs better in high accuracy index because SuperPoint and SuperGlue give better feature point matches than HFNet and KNN. Our method also outperforms ORB-SLAM3 because VLS eliminates global drift.

In OpenLORIS dataset, our method has higher success rate since train and test sets are partially overlapped. Illumination and scene changes also affect localization accuracy. In some cases, our method fails to balance between VLS and VIO observation and has larger error than ORB-SLAM3.

\subsection{Runtime Evaluation}
Average runtime of VLS and fusion are 215.21ms and 21.58ms separately. VLS is only for key frames and our system will throw key frames when their timestamp falls behind. Overall, our system is real-time.

\section{CONCLUSIONS}

In this work we present a loosely-coupled visual localization architecture. Our approach relies on smart schedule strategy to realize real-time localization frequency by balancing the computation complexity between the client and server. Fusion of global localization and VIO supply the mutual assistance to overcome some corner cases. Besides, some improvements of individual modules, including observation constraints, PGO and service scheduling, ensure the high localization performance. We provide sufficient statistics to prove the robustness, precision and speed under diverse conditions of our work on 4Seasons and OpenLORIS datasets, which outperforms some state-of-the-art visual SLAM solutions. In the future, we will focus on higher precision under degenerate conditions for VLS. Generally, tightly-coupled SLAM solutions possess promising performances in accuracy and stability. Therefore we look forward to transfering the whole architecture into a tightly-coupled one.







\bibliographystyle{ieeetr}
\bibliography{vinps_bib}

\end{document}